\documentclass[10pt,twocolumn,letterpaper]{article}

\usepackage{cvpr}
\usepackage{times}
\usepackage{epsfig}
\usepackage{graphicx}
\usepackage{amsmath}
\usepackage{amssymb}
\usepackage{CJKutf8}
\usepackage{multirow}

\usepackage[breaklinks=true,bookmarks=false]{hyperref}

\cvprfinalcopy 

\def\cvprPaperID{****} 

\begin{document}
	\graphicspath{{image/}}
	\title{Vehicle Re-identification in Aerial Imagery: Dataset and Approach}
	
	\author{Peng Wang\footnotemark[2]{ }, Bingliang Jiao\footnotemark[2]{ }, Lu Yang\footnotemark[2]{ }, Yifei Yang\footnotemark[2]{ },  Shizhou Zhang\footnotemark[2]{ },  Wei Wei\footnotemark[2]{ }, Yanning Zhang\footnotemark[2]{ }\\
		School of Computer Science and Engineering, Northwestern Polytechnical University, Xi’an, China\footnotemark[2]\\
}
\maketitle

	
\begin{abstract}
In this work, we construct a large-scale dataset for vehicle re-identification (ReID), 
which contains $137$k images of $13$k vehicle instances captured by UAV-mounted cameras.
To our knowledge, it is the largest UAV-based vehicle ReID dataset. 
To increase intra-class variation, 
each vehicle is captured by at least two UAVs at different locations, with diverse view-angles and flight-altitudes.
We manually label a variety of vehicle attributes, including vehicle type, color, skylight, bumper, spare tire and luggage rack.
Furthermore, for each vehicle image, the annotator is also required to mark the discriminative parts that helps them to
distinguish this particular vehicle from others.
Besides the dataset, we also design a specific vehicle ReID algorithm to make full use of the rich annotation information.
It is capable of explicitly detecting discriminative parts for each specific vehicle 
and significantly outperforms the evaluated baselines and state-of-the-art vehicle ReID approaches.

\end{abstract}	
	
\section{Introduction}
\begin{figure}[t]
	\begin{center}
		\includegraphics[width=1\linewidth]{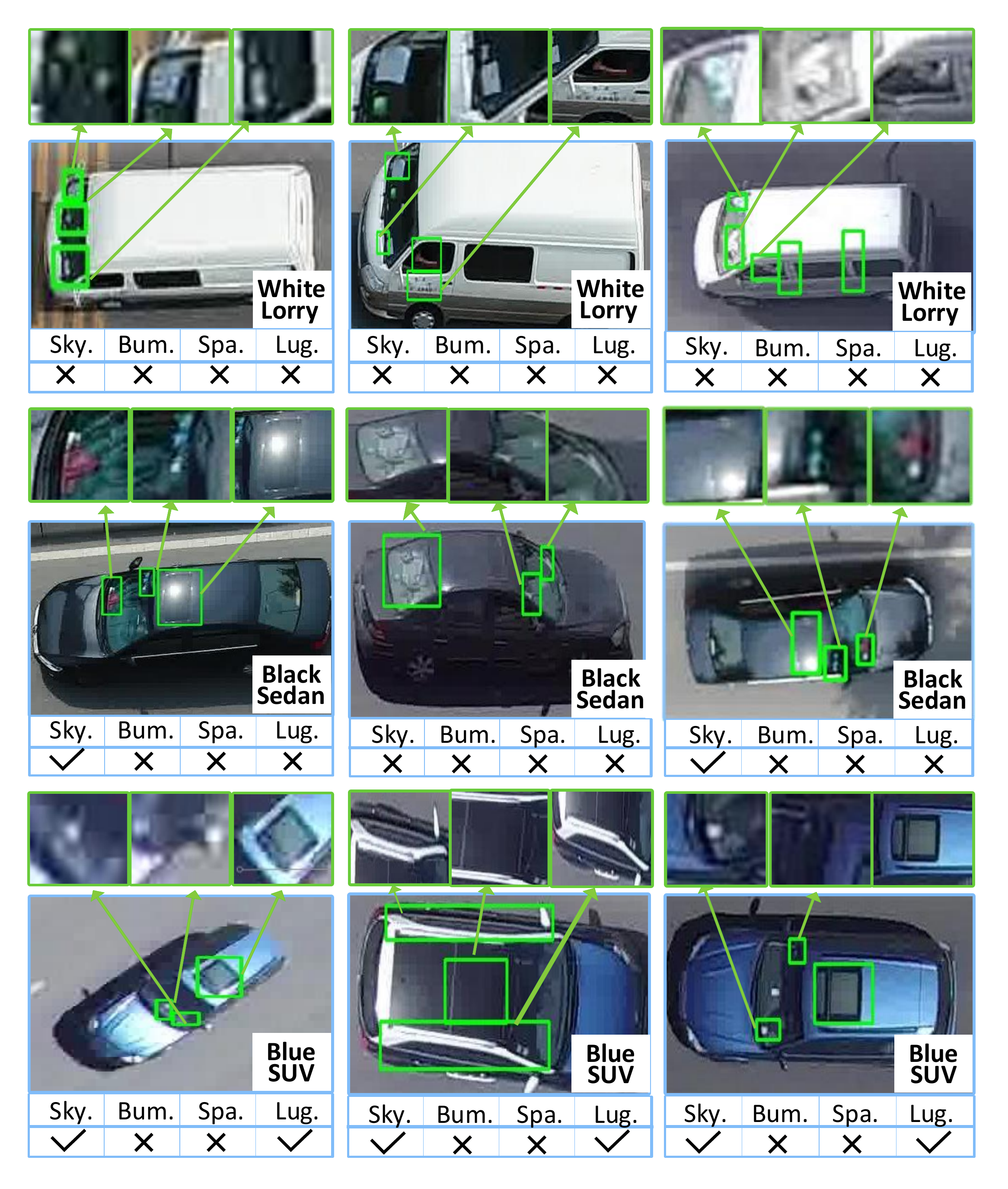}
	\end{center}
    	\caption{
    	Illustration of our collected dataset for UAV-based vehicle ReID.
    	Can you figure out which two images belong to the same vehicle instance in each row? The answer is at the end of our paper.
    	To aid in-depth research, a rich set of information is annotated in our dataset, including color, vehicle type, Skylight (Sky.), Bumper (Bum.), Spare tire (Spa.), Luggage rack (Lug.) and discriminative parts.
    	}
	\label{fig:Overview}
\end{figure}

With the rapid development of the Unmanned Aerial Vehicles (UAVs), 
the UAV-based vision applications have been drawing an increasing attentions from both industry and academia~\cite{zhuvisdrone2018}.
Existing UAV-related research and datasets in computer vision are mainly focused on the tasks of 
object detection~\cite{zhou2018scale,zhang2018w2f,zhou2018rich}, 
single or multiple object tracking~\cite{mueller2018a,bhat2018unveiling,zhu2018distractor,song2018visual,sun2018learning,song2017convolutional}, 
action recognition~\cite{barekatain2017okutama,perera2018uav,singh2018eye}  
and event recognition~\cite{oh2011large}.
However, the UAV-based vehicle re-identification is rarely studied, although
it has a variety of potential applications such as long-term tracking, visual object retrieval, \etc. 
One of the reasons is the lack of the corresponding publicly available dataset, 
which will take a large amount of human efforts for UAV flying, video capture and data annotation.
Existing vehicle ReID datasets~\cite{yang2015a,liu2016deep,liu2016large} are collected by fixed surveillance cameras,
which differs from UAV-mounted cameras in the view-angles and image qualities.
\begin{figure*}[t]
	\begin{center}
		\includegraphics[width=0.99\linewidth]{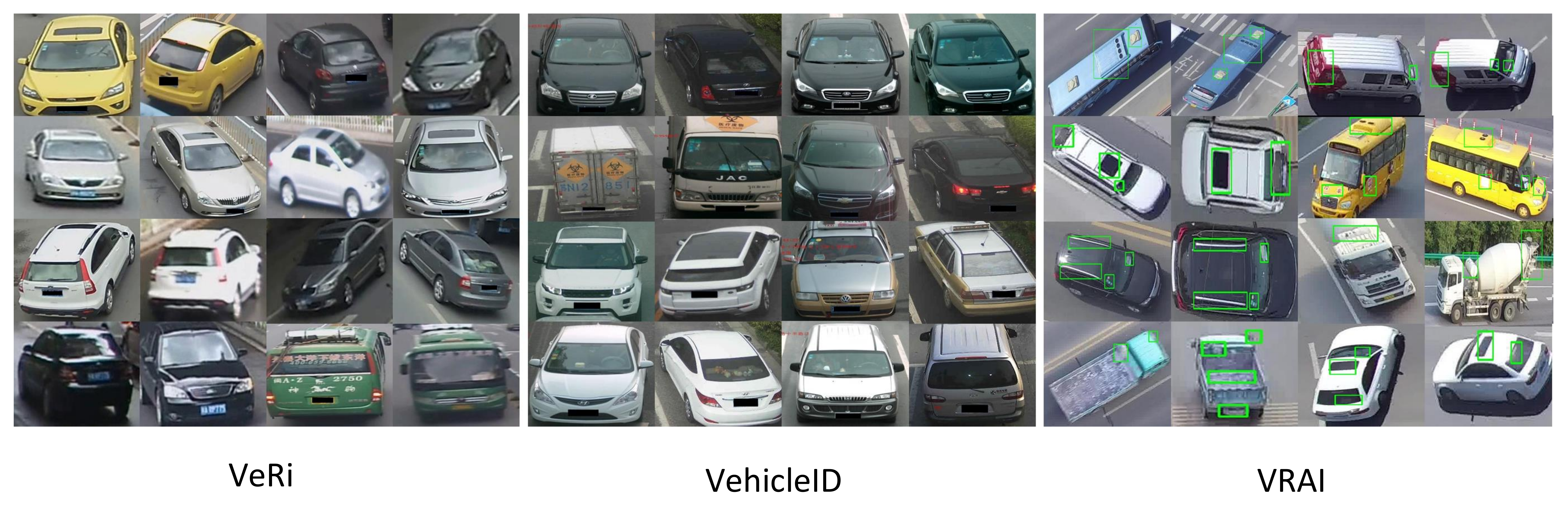}
	\end{center}
	\vspace{-4mm}
	\caption{Comparison of our VRAI dataset with other vehicle ReID datasets. Our dataset not only provides a diversity range of view-angles, but also additional annotations of discriminative parts for each particular vehicle instance. The annotated discriminative parts offer strong supervision information for fine-grained recognition.}
	\label{fig:Feature}
\end{figure*}



In this paper, we construct a large-scale vehicle ReID dataset for UAV-based intelligent applications, 
named Vehicle Re-identification for Aerial Image (VRAI). 
The VRAI dataset consists of {$137,613$} images of {$13,022$} vehicle instances.
The images of each vehicle instance are captured by cameras of two DJI consumer UAVs at different locations,
with a variety of view-angles and flight-altitudes ($15$m to $80$m).
As shown in Figure~\ref{fig:Overview}, the task of UAV-based vehicle ReID is typically more challenging than the counterpart based on fixed surveillance cameras, as vehicles in UAV-captured images are featured in larger pose variation and wider range of resolution.      

To support in-depth research, we collect a rich set of annotations,
including vehicle IDs, vehicle attributes and discriminative parts.
Images of the same vehicle instance are manually assigned with a unique ID,
according to the appearance similarity and time dependency. 
For every image, we also annotate the color ($9$ classes), 
vehicle type ($7$ classes), and whether or not having skylight, bumper, spare tire and luggage rack. 
Moreover, differing from the tasks of vehicle detection~\cite{dinesh2018carFusion}, tracking~\cite{matei2011vehicle,xiang2014monocular} and classification~\cite{zhang2012three,lin2014jointly},
vehicle re-identification relies more on small regions containing fine-grained discriminative information.
To this end, we also ask annotators to mark these discriminative parts using bounding boxes for each image in our dataset.
Figure~\ref{fig:Feature} illustrates some examples of annotated discriminative parts. We can see that many discriminative parts correspond to interior decoration of frontal window, skylight, bumper, and so on.

To summarize, the featured properties of our VRAI dataset include:

  \noindent {\bf Largest vehicle ReID dataset to date.} It contains over {$137,613$} images of {$13,022$} vehicles which is the largest UAV-based vehicle ReID dataset to our knowledge. Each vehicle has over $10$ images in average.
  
  \noindent{\bf Rich annotations.} Besides unique IDs, we also annotate the color, vehcile type, attributes \eg whether it contains skylight, spare tire \etc and discriminative parts of the images in the dataset. 
  
  \noindent{\bf Diverse view-angles and poses.} The images are taken by two moving UAVs in real urban scenarios, and the flight-altitude ranges from $15$m to $80$m. 
  It results in a large diversity of view-angles and pose variations,
  and so increases the difficulty of the corresponding ReID task.

%

Based on the rich annotation information of our dataset, we propose a novel approach for vehicle ReID from aerial images, which is capable of explicitly detecting discriminative parts for each specific vehicle and significantly outperforms other compared algorithms.




\section{Related Work}\label{sec_RW}

In this section, we briefly review the related works from the following three aspects.

\noindent\textbf{Vehicle Image Datasets.}
Recently, more and more vehicle related datasets have been collected for many research fields. 
Yang \etal~\cite{yang2015a} provide a large scale vehicle dataset, named CompCars, for fine-grained categorization and verification.
The KITTI dataset~\cite{Geiger2012CVPR} is collected to serve as a benchmark dataset for the fundamental tasks of object detection, tracking, semantic segmentation \etc.
Several vehicle ReID datasets have also been constructed. Liu \etal~\cite{liu2016large} construct a relatively small vehicle ReID dataset named VeRi which includes $40,000$ bounding boxes of $619$ vehicles. VehicleID~\cite{liu2016deep} is a much larger vehicle ReID dataset with $221,763$ images of $26,267$ vehicles in total. 
We can see from Figure~\ref{fig:Feature} that, both the VeRi and VehicleID datasets contain limited view-angles and vehicle poses, compared with our presented dataset. 

\noindent\textbf{Aerial Visual Datasets.}
With the rapid development of the commercial UAV, more and more aerial visual datasets have been constructed to facilitate the research of aerial vision tasks.
DOTA~\cite{xia2018a} and NWPU VHR-10~\cite{cheng2018Learning} are the datasets collected for object detection in aerial images which are taken by UAVs from a relatively high flying altitude. 
UAV123~\cite{mueller2018a} is a video dataset aimed to serve as a target tracking dataset, which is taken by an UAV with a relatively low flying altitude.
The Visdrone2018~\cite{zhuvisdrone2018} dataset is collected to serve as a benchmark dataset for the challenge of ``Vision Meets Drones''. 
The main tasks of the challenge are still object detection and tracking. 


\noindent\textbf{ReID Approaches.}
Person and vehicle are two important object classes in urban surveillance scenarios.
Person ReID has been very attractive in recent years.
For instance, Wei \etal~\cite{wei2018person} adopt GAN to bridge
the domain gap between different person Re-ID datasets.
Liu \etal~\cite{liu2018pose} propose a pose transferrable
person ReID framework which utilizes pose transferred
sample augmentations to enhance ReID model training.
Li \etal~\cite{li2018diversity} incorporate a multiple spatial attention model to learn the latent representations of face, torso and other body parts to improve the performance of the model. Dai \etal~\cite{dai2018cross} improve the ReID performance between infrared and RGB images by adopting a novel cross-modality generative adversarial network (termed cmGAN). Shen \etal~\cite{shen2018per} provide more precise fusion information by proposing deep similarity-guided graph neural network (SGGNN) and utilizing the relationship between probe-gallery pairs. Bak \etal~\cite{bak2018dom} alleviate the diversity in lighting conditions by introduced a new synthetic dataset and propose a novel domain adaptation technique. Ge \etal~\cite{ge2018fd} adopt a Feature Distilling Generative Adversarial Network (FD-GAN) for learning identity-related and pose-unrelated representations. There are also many other researches in this field~\cite{wang2018person,kalayeh2018human,xu2018attention,chang2018multi,li2018harmonious,guo2018efficient}.


Vehicle ReID also gains increasing attentions recently. 
For example, Wang \etal~\cite{wang2018orientation} propose to utilize an orientation invariant feature embedding module and a spatial-temporal regularization module
to improve the vehicle ReID performance.
Shen \etal~\cite{shen2017learning} propose a two-stage framework which incorporates complex spatio-temporal information for effectively regularizing the
re-identification results.
MGN~\cite{wang2018learning} uses the first three layers of Resnet50 to extract shared image features, and relies on three independent branches to extract the high-level semantic features.
The RNN-HA presented by Xiu~\etal in ~\cite{xiu2018coarse} consists of three interconnected modules. The first module creates a representation of the vehicle image, the second layer module models the hierarchical dependencies, and the last attention module focuses on the subtle visual information distinguishing specific vehicles from each other.
RAM~\cite{liu2018ram} extracts local features to assist in the extraction of global features. 



\section{Dataset}\label{sec_DT}
In this section, we give details of the 
constructed
VRAI dataset, including the hardware sensors, collecting process, and annotations.
\subsection{Data Collection}

We use two DJI Phantom$4$ UAVs to simultaneously 
shoot
videos at two adjacent locations (in total we select $11$ location pairs), 
in order to capture images of individual vehicles with different view-angles and context.
The two UAVs are controlled to have no overlaps in visible areas. 
To increase the diversity of object resolutions, two UAVs are kept in different altitudes, ranging from $15$m to $80$m.
In the process of controlling UAV, we 
adopt
various sport modes such as hovering, cruising, and rotating to collect data undergoing viewpoint and scale changes.

With more than $200$ man-hours of UAV flying and video shooting, 
we finally collect $350$ pairs of video clips, with a total length of $34$ hours (approximately $3$ minutes each clip).
In each clip, we sample frames at every $0.5$ second and obtain $25,200$ images in total. 

\begin{figure}[t]
	\begin{center}
		\includegraphics[width=1\linewidth]{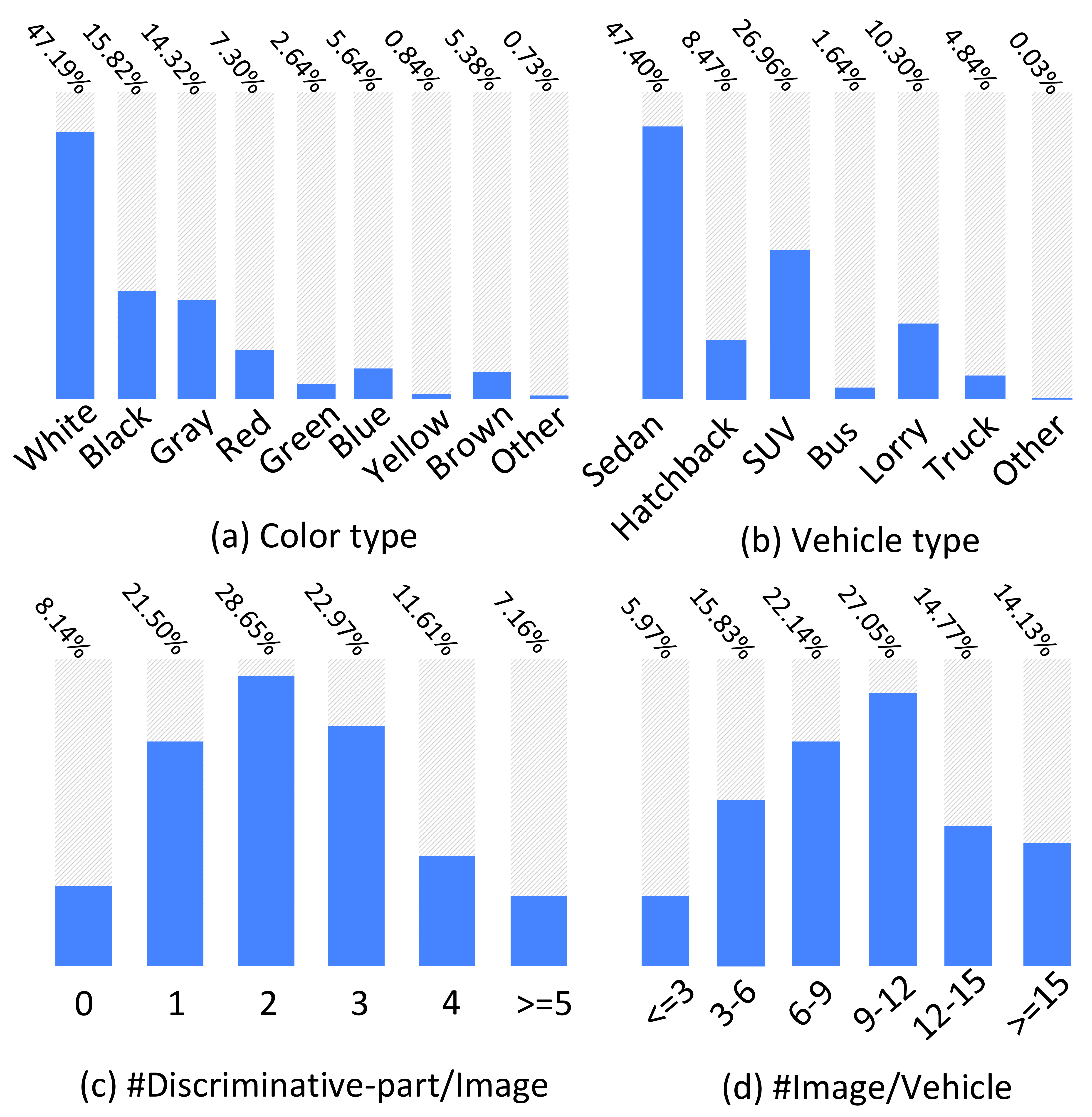}
	\end{center}
	\caption{The statistical information about (a) color; (b) Vehicle type; (c) discriminative part number per image; (d) image number per vehicle. White, black and gray are the mainstream colors. Sedan, SUV and lorry are more popular than other vehicle types. About $91.8\%$ instances are labeled with at least one discriminative part. And $94.0\%$ vehicles have more than $3$ images.
}
	\label{fig:C&T}
\end{figure}
\subsection{Annotation}
We develop 
a
software to perform the following four steps of annotation:   

{\bf1)~Object Bounding Box Annotation:}
In each image, four corners of all visible vehicles are manually marked, and the smallest rectangular bounding box containing all the four corners is automatically calculated and stored.
The distribution of bounding box resolution is demonstrated in Figure~\ref{fig:re}, with $42.90\%$ not larger than $50$k pixels and 
$8.8\%$ larger
than $200$k pixels. 
Besides, we can clearly see that the image resolution of our dataset changes more than that of VehicleID dataset.
We use $1000$ man-hours to finish this step of annotation.
%
%

{\bf2)~Cross-Camera Vehicle Matching:}
The most time consuming annotation step is cross-camera object matching, 
in which instances of the same vehicle appearing in two video clips need to be grouped together. 
As license plates are not visible from aerial images, annotators can only rely on the appearance similarity and temporal correspondence.  
After spending around $2500$ man-hours, we collect $137,613$ instances (bounding boxes) of $13,022$ individual vehicles (IDs).
From Figure~\ref{fig:C&T} (d), we can find that $94.0\%$ vehicles have more than $3$ annotated bounding boxes taken by two cameras. 
%
%
%
%

{\bf3)~Attributes Categorization:}
In this step, each of $137613$ matched instances is manually labeled with several attributes, 
including color (White, Black, Gray, Red, Green, Blue, Yellow, Brown, and Other), 
vehicle type (Sedan, Hatchback, SUV, Bus, Lorry, Truck and Other)
 and four binary attributes (if containing Skylight, Spare Tire, Bumper or Luggage Rack).
This annotation step takes in total $62$ man-hours.
The distributions of annotated color and vehicle type are shown in Figure~\ref{fig:C&T}.
We can find that While, Black and Gray are the dominating colors, 
while
Sedan, SUV and Hatchback are the dominating vehicle types.  
%
%
%
%
%

{\bf4)~Discriminative Parts Annotation:}
Distinguishing vehicles with similar attributes 
requires
fine-grained information for each specific ID. 
For each bounding box, we manually annotate multiple discriminative parts with small bounding boxes.
These parts are considered by annotators to be crucial for distinguishing a particular instance from others.
As shown in Figure~\ref{fig:Overview}, a large number of annotated discriminative parts are frontal windows, luggage racks, skylights, and headlights.
We also allow annotators to skip this step, if he/she cannot find any part that is discriminative. 
From Figure~\ref{fig:C&T} (c), we can find that $91.8\%$ instances are labeled with at least one discriminative part, and $63.2\%$ instances come with $2$ to $4$ annotated parts.
This step takes $1300$ man-hours of annotation.

\subsection{Comparison with Other Datasets}
In Table~\ref{tab:compare}, our dataset is compared with existing datasets for vehicle re-identification or fine-grained recognition,
in terms of instance number, image number, attributes annotation and discriminative parts annotation.
The difference between our dataset and other datasets is summarized as follows. 

{\bf1)~Capturing Platform:}  
To our best knowledge, the proposed dataset is the first one for vehicle ReID in aerial images. The images in our dataset are captured from a diverse set of view-angles by cameras mounted on moving UAVs, while images in other vehicle datasets are captured by fixed cameras. In addition, the images in our dataset are shot by two UAVs controlled by different pilots.

{\bf2)~Data Size:}
Our dataset contains an order of magnitude larger number of instances than CompCars~\cite{yang2015a} and VeRi~\cite{liu2016large}. Besides, in our dataset each instance contains over $10$ images on average, while VehicleID~\cite{liu2016deep} only contains $8$ images. Our dataset has certain advantages in the richness of the counts of the training data.

{\bf3)~Data Annotations:} 
Among the listed datasets, ours is the only one equipped with discriminative part annotations, which offers strong supervision information for fine-grained recognition. Besides, VRAI has also collected color, type and other attributes annotations.

{\bf4)~View-angle Variation:} 
Since our dataset is collected by UAVs, the view-angles of the shot images change frequently. Compared with VehicleID~\cite{liu2016deep} which collected by fixed cameras, our dataset contains a more diverse range of view-angles, as shown in Figure~\ref{fig:Feature}.

\begin{figure}[t]
	\begin{center}
		\includegraphics[width=1\linewidth]{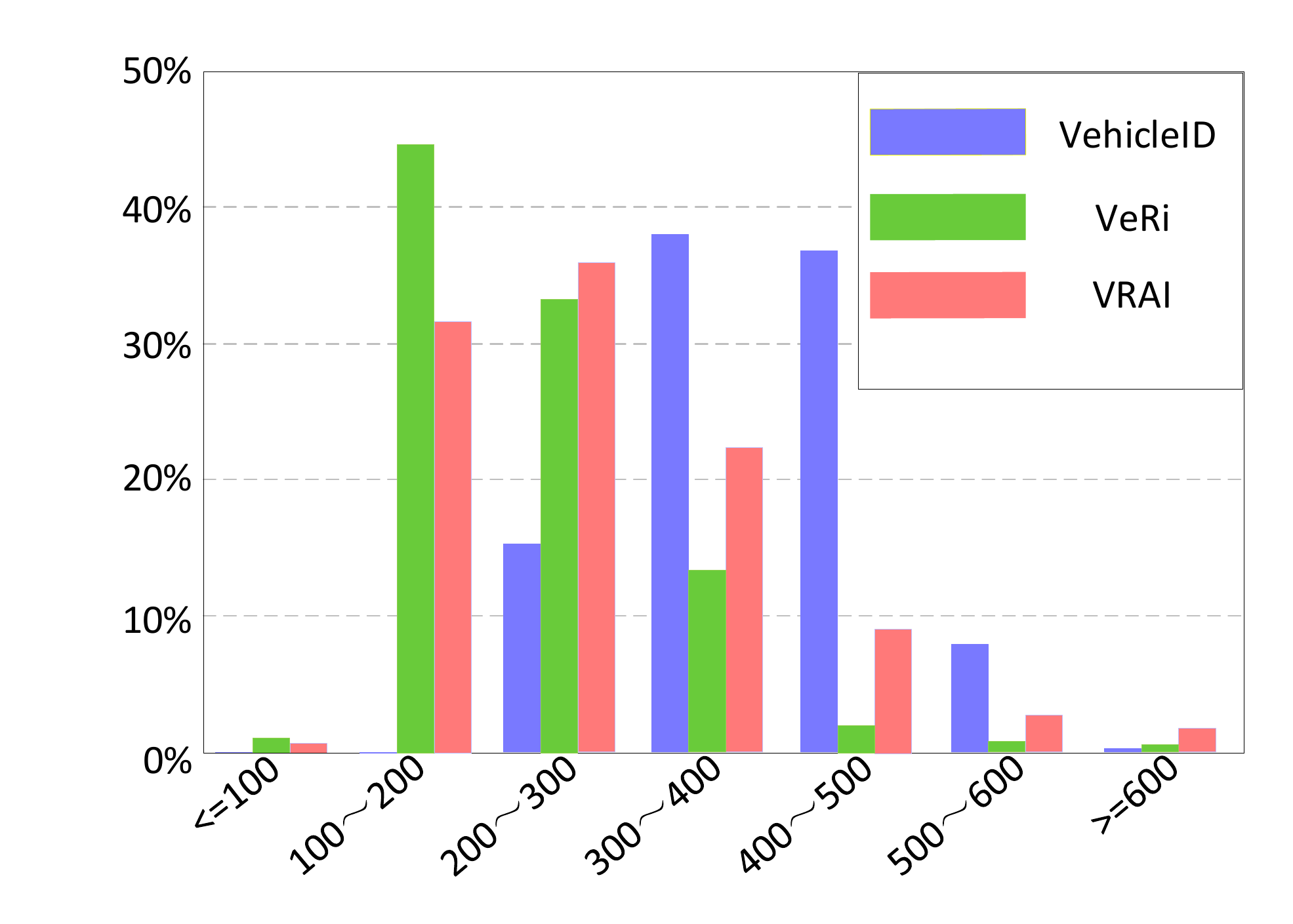}
	\end{center}
	\caption{ The image resolution distribution of VehicleID, VeRi and VRAI. From the figure we can see that VRAI has a wider resolution distribution.}
	\label{fig:re}
\end{figure}

{\bf5)~Image Resolution:} 
During the data collect process, the resolution of the shot image fluctuates due to the change of the altitude of the UAVs. Under these circumstances, the image resolution of our dataset changes more than that of VehicleID dataset, which is shown in the Figure~\ref{fig:re}.

%
\begin{figure*}[t]
    \begin{center}
        \includegraphics[width=1\linewidth]{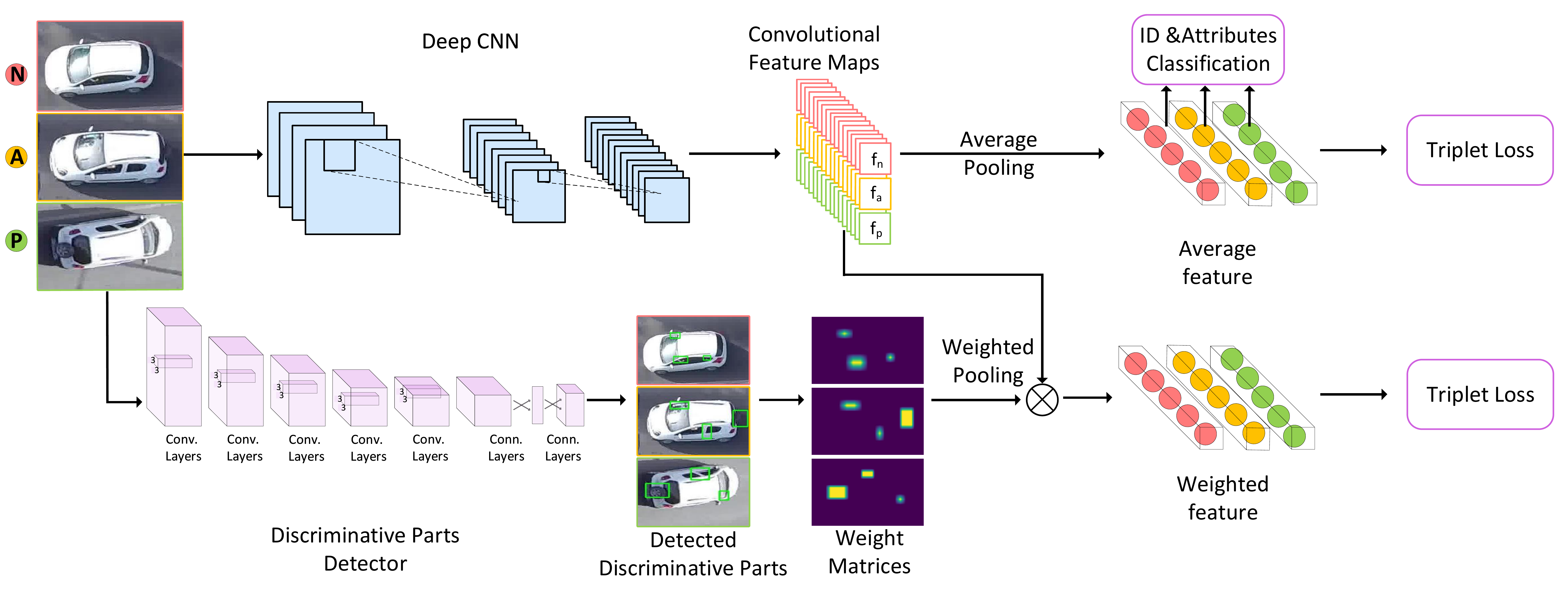}
    \end{center}
    \caption{The overall structure of our proposed model. 
    Our model has two main components, one is a multi-task branch which trains model through a series of classification losses and a triplet loss using the average feature. 
    The triplet loss is trained with the training triplets, \ie, $\langle$Anchor (A), Positive (P), Negative (N)$\rangle$s.
    The other branch is a Discriminative Part Detection module which uses weight matrix for weighted pooling. And the Discriminative Part Detection module uses weighted feature to train model. }
    \label{fig:structure}
\end{figure*}

\begin{table}
	\begin{center}
	\scalebox{0.85}{
		\begin{tabular}{l|c|c|c|c}
			\hline
			Dataset/Char. & $\#$ ID & $\#$ Image & Attr. & Dis. parts \\
			\hline
			CompCars~\cite{yang2015a} & $1716$ & $208826$ & $\surd$  & $\times$\\
			VehicleID~\cite{liu2016deep} & $26267$ & $221763$ & $\surd$  & $\times$\\
			VeRi~\cite{liu2016large} & $619$ & $40215$ & $\times$  & $\times$\\
			VRAI & $13022$ & $137613$ & $\surd$  & $\surd$\\
			\hline
		\end{tabular}}
	\end{center}
	\caption{The ``char." represents characters, the ``attr." represents attributes and the ``Dis." represents discriminative. Note that CompCars only contains the model annotation of each vehicle, and only $30$\% of the VehicleID images are marked with attribute information. Our dataset is the only one providing discriminative part annotations.}
	\label{tab:compare}
\end{table}


\section{Approach}\label{sec_method}
In this section, we present a ReID algorithm based on the rich annotations of the proposed dataset, 
which is capable of explicitly detecting discriminative parts of a particular vehicle instance.  
Our ReID algorithm is detailed in the following.

\subsection{Overall Framework}
As shown in Figure \ref{fig:structure}, the main structure of the proposed method can be divided into two branches. 
The first one is a multi-objective model whose backbone network is ResNet-50~\cite{he2016deep} pretrained on ImageNet~\cite{deng2009imagenet}. 
In this branch, 
we make full usage of the rich annotations to train a multi-objective model for retrieval, ID classification and attributes classification.
In the other branch, a YOLOv2~\cite{redmon2016yolo} detector is separately trained with annotated discriminative parts.
The detected bounding boxes are used to construct a weight matrix for aggregating convolutional features of ResNet-50.
Finally, the weighted features are employed to train another retrieval model for ReID with the triplet loss.

\subsection{Multi-Objective Learning}
The convolutional feature after average pooling and batch normalization 
is trained with multiple objectives, 
including retrieval, ID classification, color, vehicle type and
attributes classification.

Several loss functions are used for different tasks respectively.
The cross entropy loss ($L_{ce}$) is adopted for multi-class problems such as ID, color and vehicle type classification.
%
%
For the multi-label attributes classification problems, we use Binary Cross Entropy loss.
%
The triplet loss function is used for the retrieval task.
%

\subsection{Discriminative Part Detection}
In order to distinguish vehicles of similar color or type, we train a detector based on YOLOv2~\cite{redmon2016yolo} using the annotations of discriminative parts. 
This detector is trained separately and all discriminative parts are treated as the same class.
Thanks to the huge number of annotated discriminative parts in our dataset ($322853$), we are able to train an effective detector. 
For example, the detector can extract many valuable discriminative parts such as skylight, window, LOGO, even if the ground truth only provides the skylight.
For each vehicle image, we extract top-$3$ bounding boxes generated by the discriminative parts detector.


%

\subsection{Weighted Feature Aggregation}
In addition to average pooling, we also extract a weighted feature based on the detected discriminative parts.
The backbone network in our model is ResNet-50, with the input size of $352\times352$ and the output feature map size of $2048\times11\times11$.
An $11\times11$ weight matrix is generated by increasing the weights of pixels inside discriminative parts: 
\begin{equation}
\mbox{weight}_{i}=\begin{cases}~\gamma\qquad i\in \mathcal{D},\\
~1\qquad \mbox{otherwise},\end{cases}
\label{eq:gamma}
\end{equation}
where $i$ denotes a pixel index; $\gamma$ is a predefined scalar larger than $1$; $\mathcal{D}$ denotes the region of detected discriminative parts.
With this weight matrix, we perform a weighted pooling over the feature map of size $2048\times11\times11$, rendering a weighted feature of size $2048$.
	
\section{Experiments}\label{sec_expr}
In this section, we show the experimental results of the proposed vehicle ReID approach on our VARI dataset, 
including attribute classification, discriminative part detection, ablation studies, comparison experiments with baseline and state-of-the-art vehicle ReID methods.  
We also perform a human performance evaluation to measure the potential of our dataset.
To clarify, firstly we give the evaluation protocols and implementation details.

\subsection{Evaluation Protocols and Implementation}
The VARI dataset is split to the training set and test set, among which the training set contains $66,113$ images with $6,302$ IDs, and the test set contains $71,500$ images with $6,720$ IDs.
The test set is further divided into a query set ($25\%$ images) and a gallery set ($75\%$ images). 
Meanwhile, for each image in the query set,
it is ensured that at least one image in the gallery has the same ID with that query while being captured by different cameras. 
Consequently, there are $15,747$ images in the query set and $55,753$ images in the gallery set. 
At the test stage, each vehicle image in the query set is used to retrieve the same vehicle in the gallery set. 
As for evaluation criteria, we adopt the popular mean Average Precision (mAP) and Cumulative Matching Cure (CMC) as in other ReID works.



During experiments, we use Resnet-50 pre-trained on ImageNet as backbone. 
Each image is resized into $352\times352$. 
In the training phase, we rotate each image with $90$, $180$ or $270$ degrees clockwisely with a probability of $0.2$, 
and flip it horizontally with a probability of $0.5$. 
The margin of the triplet loss~\cite{liu2017end} is set to $0.3$, 
and the mini-batch size is set to $72=18\times 4$, with $18$ identities and $4$ images for each ID. 
We use the Adam optimizer with an initial learning rate of $10^{-3}$ 
and the learning rate starts to decay from the $151$th epoch. 
All models are trained with $300$ epochs. 
To improve the performance, we train the model using the BatchHard Triplet Loss~\cite{hermans2017in} and the ID classification loss jointly.
Two NVIDIA $1080$Ti GPUs are used for model training.

\begin{table}
	\begin{center}
		\resizebox{0.35\textwidth}{!}{
		\begin{tabular}{|l|c|c|c|}
			\hline
			\multicolumn{2}{|c|}{} & Accuracy (\%) \\
			\hline
			\multicolumn{2}{|c|}{{Color}} & $86.25$\\
			\multicolumn{2}{|c|}{{Type}} & $81.88$\\
			\hline
			\multirow{4}{*}{{Attributes}} &Skylight & $90.16$\\
			&{Bumper} & $82.44$\\
			&{Spare Tire} & $97.35$\\
			&{Luggage Rack} & $85.67$\\
			\hline
		\end{tabular}}
	\end{center}
	\caption{The classification results of our model for color, vehicle type, Skylight, Bumper, Spare tire and Luggage Rack. 
	It indicates that skylight and spare tire are more distinguishable than others, and the vehicle type and bumper are more difficult to classify.}
	\label{tab:Multi}
\end{table}

\subsection{Attributes Classification}
\begin{figure}[t]
	\begin{center}
		\includegraphics[width=0.80\linewidth]{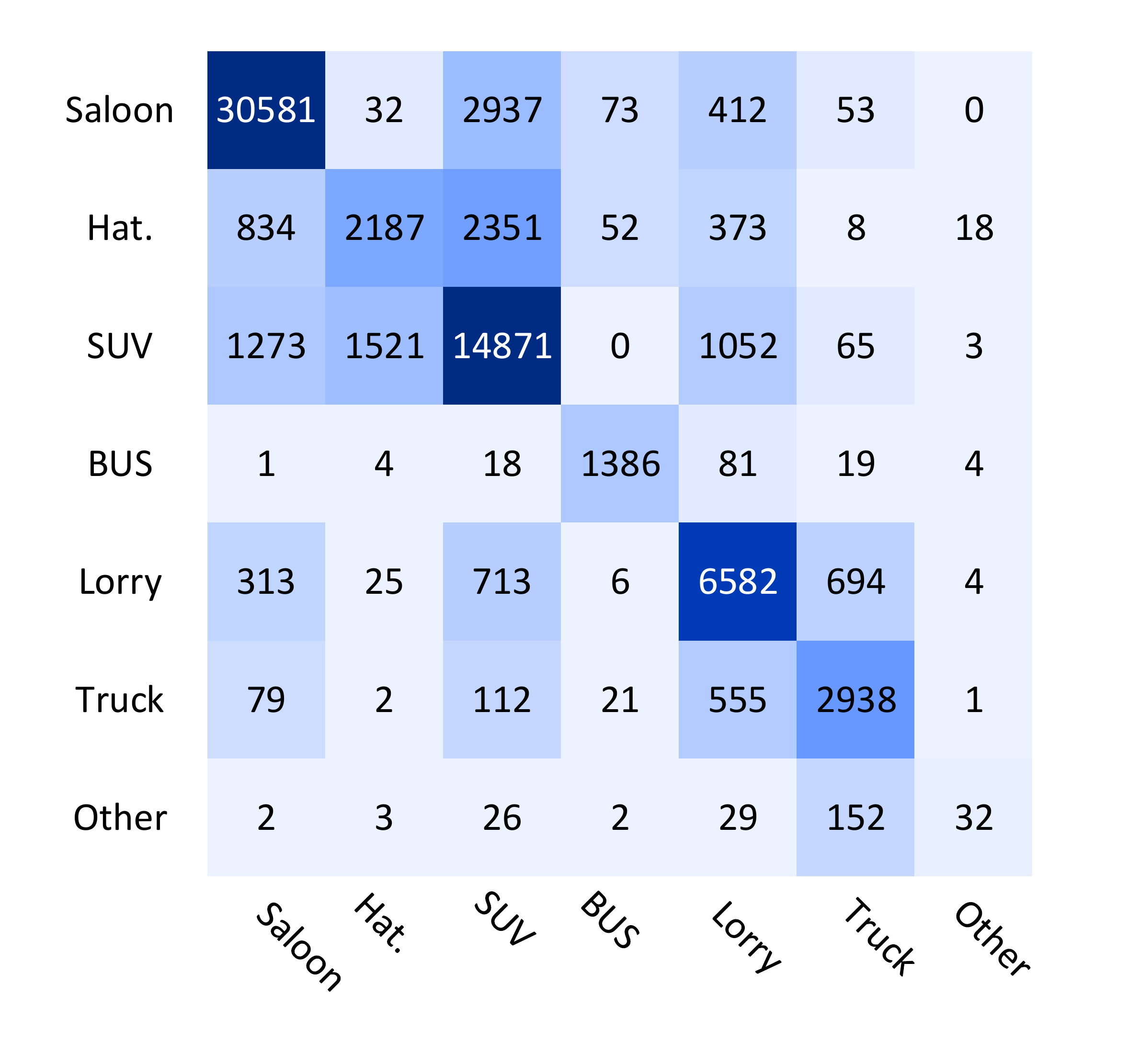}
	\end{center}
	\vspace{-5mm}
	\caption{
	The confusion matrix of vehicle type classification.
	We can see that the most classification errors are incorrect identification between Hatchbacks and SUVs.
	These errors are caused by the visual similarity between these two vehicle types, given that actual vehicle sizes are somewhat difficult to estimate.}
	\label{fig:matrix}
\end{figure}

The results of color, vehicle type and attributes classification of our model are shown in Table~\ref{tab:Multi}.
It can be seen that all the evaluated accuracies are over $82\%$, except for that of vehicle type.
To further analyze the classification result of vehicle type, the confusion matrix is illustrated in Figure~\ref{fig:matrix}. 
We can find that Hatchback and SUV are the two classes mostly confusing our classification model.  
And the reason is probably that Hatchback and SUV are indeed visually similar from top-view, without taking vehicle size into account. 

\begin{figure}[t]
	\begin{center}
		\includegraphics[width=1\linewidth]{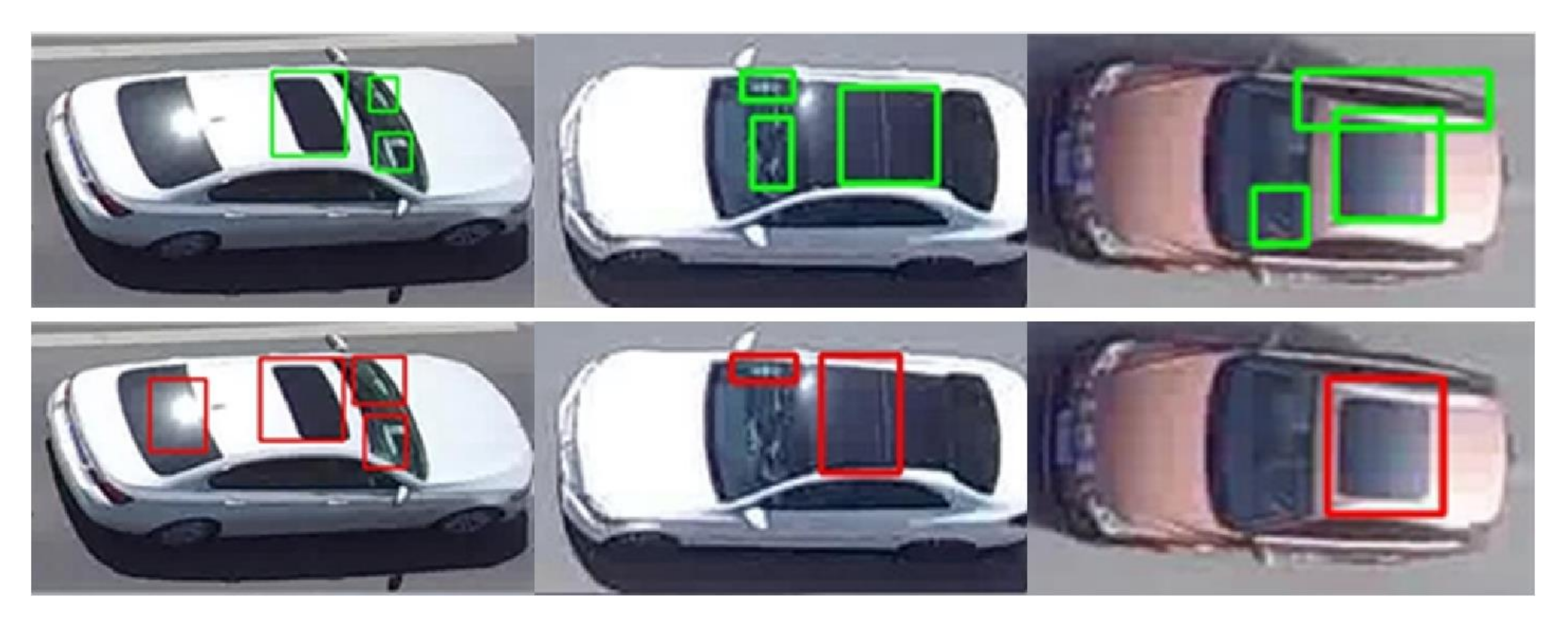}
	\end{center}
	\caption{Sample results of discriminative part detection. Red and green bounding boxes represent ground truth and predicted results respectively. Only the top$3$ bounding boxes are depicted in the predicted images. We can clearly find that our model has a good performance so that there is little difference between the predicted results and the ground truth.}
	\label{fig:Exm}
\end{figure}

\subsection{Discriminative Part Detection}
For the experiment of discriminative part detection, the 
pre-trained darknet is used to fine-tune YOLOv2~\cite{redmon2016yolo} model on VRAI. 
For simiplicity, all discriminative parts are considered as one class.
The multi-scale training scheme is adopted where the resolution of the input images is resized ranging from $320$ to $608$.
The learning rate is set to $0.001$ and then decreased  by a factor of $10$ at the $10$th and $15$th epochs.
The mini-batch  size is set to $72$.
The momentum is set to $0.9$ and the weight decay rate is $0.0005$.

In the testing phase, we respectively set the confidence threshold, NMS threshold, and IOU threshold to $0.25$,  $0.4$, and $0.5$ empirically, 
and achieve a result of precision=$44.07\%$, recall=$49.48\%$ and F-score=$46.62\%$. 
Fig.~\ref{fig:Exm} showcases the detection results and the ground truth of the discriminative parts. 
\begin{table*}
	\begin{center}
		\resizebox{0.9\textwidth}{!}{ 
			\begin{tabular}{l|c|c|c|c|c|c|c}
				\hline
				& Backbone  & Attribute  & D.P. & mAP (\%) & CMC-1 (\%)& CMC-5 (\%)& CMC-10 (\%)\\
				& Model  & Annotation  & Annotation &   &  &   &  \\
				\hline
				MGN~\cite{wang2018learning} &Resnet-50& &  & $69.49$ & $67.84$& $82.83$ & $89.61$ \\
				RAM~\cite{liu2018ram} & Resnet-50 & &  & $69.37$ &$68.58$& $82.32$ & $89.88$ \\
				RAM~\cite{liu2018ram} & VGG-16 & &  & $57.33$ & $72.05$ & $81.62$ & $56.82$ \\
				RNN-HA~\cite{xiu2018coarse} & Resnet-50 & $\surd$ & & $74.52$ & $77.43$& $87.38$ & $92.65$ \\
				\hline
				Triplet loss  & Resnet-50 & & &  $46.99$ & $50.64$ & $71.49$ & $80.40$ \\
				Contrastive loss & Resnet-50 &   &  & $48.23$ & $52.23$ & $72.28$ & $81.29$\\
				ID classification Loss & Resnet-50    &   &  & $72.96$ & $75.96$ & $87.01$ & $92.70$ \\
				Triplet + ID Loss & Resnet-50   &  & & $77.28$& $79.13$ & $88.47$ & $93.64$ \\
				Triplet + ID Loss & Resnet-101   &  & & $77.48$& $79.59$ & $88.31$ & $93.69$ \\
				Triplet + ID Loss & Resnet-152   &  & & $77.54$& $79.33$ & $88.18$ & $93.47$ \\				
				\hline
			    Ours Multi-task & Resnet-50 & $\surd$  &   & $78.09$ & $79.83$ & ${\textbf{89.05}}$ & $94.09$ \\
				Ours Multi-task + DP & Resnet-50   & $\surd$ & $\surd$& ${\textbf{78.63}}$ & ${\textbf{80.30}}$ & $88.49$ & ${\textbf{94.45}}$ \\
				\hline
		\end{tabular}}
	\end{center}
	\caption{Our final model is compared with other baselines and intermediate models. 
		We can find that the ID classification and weighted feature have greater contribution for improving performance.}
	\label{tab:results}
\end{table*}

\begin{table}
	\begin{center}
	\small
		\resizebox{0.48\textwidth}{!}{
		\begin{tabular}{l|c|c|c|c}
			\hline
			Feature tested & mAP ($\%$) & CMC-1 ($\%$) & CMC-5 ($\%$) & CMC-10 ($\%$)\\
			\hline
			Avg. Feat. & $78.31$ & $80.05$ & $89.01$ & $94.22$\\
			Weight. Feat. & $\mathbf{78.63}$ & $\mathbf{80.30}$ & $\mathbf{88.49}$ & $\mathbf{94.45}$\\
			\hline
		\end{tabular}}
	\end{center}
	\caption{The comparison of our {\bf Multi-task + DP} model using average feature and weighted feature respectively for distance calculation. The results show that the performance of the model using weighted feature is better than the one using average feature.}
	\label{tab:Compare}
\end{table}

\begin{table}[]
\begin{center}
\scalebox{0.85}{
\begin{tabular}{c|c|c}
				\hline
				Weighting Parameter $\gamma$  & mAP ($\%$) & CMC-1 ($\%$)\\
				\hline
				$1.1$ & $78.34$ & $80.07$ \\
				$1.3$ & ${\textbf{78.63}}$ & ${\textbf{80.30}}$ \\	
				$1.5$ & $78.48$ & $80.19$  \\
				$1.7$ & $78.36$ & $80.06$  \\
				$1.9$ & $78.41$ & $80.09$ \\	
				\hline
		\end{tabular}
}
\end{center}
\caption{The empirical study on the weighting parameter $\gamma$ of our {\bf Multi-task + DP} model, which shows that $1.3$ yields better performance.
         Note that the {\bf Multi-task + DP} model performs better than the {\bf Multi-task} model, with all of the selections of $\gamma$, in terms of mAP and CMC-1.}
	\label{tab:abla}
\end{table}

\begin{table}[]
\begin{center}
\scalebox{0.85}{
\setlength{\tabcolsep}{1mm}{
\begin{tabular}{l|c c c c}
			\hline
			& Annotator 1 & Annotator 2 & Annotator 3 &  Ours \\
			\hline
			Accuracy & $98\%$ & $98\%$ & $96\%$& $80\%$ \\
			\hline
		\end{tabular}
		}
}
\end{center}
\caption{The performance comparison between human and our algorithm, on $100$ randomly selected queries. 
         We can see that the average human accuracy is $97\%$, which is $17\%$ higher than our algorithm.
         It shows that there is still much room for improving the algorithm performance.}
	\label{tab:hu}
\end{table}

\begin{figure}[t]
	\begin{center}
		\includegraphics[width=0.9\linewidth]{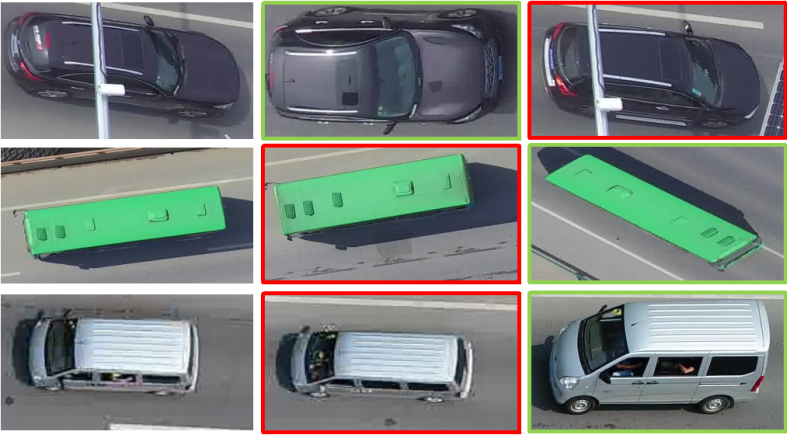}
	\end{center}
	\caption{Comparison of human performance and the performance of our algorithm. 
	The left, middle and right columns correspond to queries, the results of our model and the results of human, respectively.
	Correct and incorrect results are marked by green and red boxes respectively.
	We find that the vehicle pose and the angles of view are the main factor that causes the algorithm to generate a wrong result. For human, the error occurs when the recognition requires extremely fine-grained information.}
	\label{fig:hum}
\end{figure}
\subsection{Ablation Study}
In this subsection, to verify the effectiveness of the proposed method and 
to show how much each component of the proposed model contribute to the final performance, 
we report the vehicle ReID performance compared with several baseline methods and the ablation study on model hyperparameters. 
The detailed experimental results are shown in Table~\ref{tab:results}.
The following four methods are chosen as baselines. 
1) {\textbf{Triplet Loss.}} The model is trained with only a triplet loss on the average feature;
2) {\textbf{Constrastive Loss.}} The model is trained with only a constrastive loss on the average feature;
3) {\textbf{ID Classification Loss.}} The model is trained with a single ID classification loss;
4) {\textbf{Triplet$+$ID Loss.}} The model is jointly trained with both the triplet loss and ID classification loss on the average feature.
Without using any extra attributes annotations, the four models can be applied to any ReID data and served as baselines in our experiments.
From Table~\ref{tab:results}, it can be seen that the ID classification loss contributes more in improving the performance.
We also evaluate different CNN backbones (Resnet-50, Resnet-101 and Resnet-152) using the \textbf{Triplet$+$ID Loss} baseline.
We find that the ReID performance is slightly improved with deeper CNN models, but at a cost of higher computational burden.

Compared to {\textbf{Triplet$+$ID Loss}}, our {\bf Multi-task} model is trained with additional attribute classification losses, 
including color, vehicle type and other attributes. With the help of the attribute information, a slightly better accuracy is achieved.
As trained without the weighted feature, the baseline methods and our {\bf Multi-task} model do not rely on the discriminative part detection branch of the proposed model.
While our {\bf Multi-task + DP} model introduces an additional triplet loss on the weighted feature, and the final model consists of all the branched losses.
The results of {\bf Multi-task + DP} validate that weighted features from detected discriminative parts gives a significant improvement.

We also test the sensitivity of our {\bf Multi-task + DP} model to the weighting parameter $\gamma$ in Equation~\eqref{eq:gamma}.
As shown in Table~\ref{tab:abla}, the  {\bf Multi-task + DP} model outperforms the {\bf Multi-task} model using all the evaluated values of $\gamma$.

For our {\bf Multi-task + DP} model, we also compare using the average feature and the weighted feature for distance computation.  
It can be clearly seen from Table~\ref{tab:Compare} that weighted feature significantly outperforms the average feature, 
in terms of all the mAP, CMC-$1$, CMC-$5$ and CMC-$10$, which shows that the detected discriminative parts are indeed beneficial for recognizing individual vehicle instances. 

\subsection{Comparison with State-of-the-art Approaches}
Although there is little vehicle ReID methods specificly designed for aerial images, 
to show the superme performance of our algorithm, 
we compare the experimental results of our methods with three state-of-the-art vehicle ReID methods for ground-based scenarios. 
The three chosen methods are MGN~\cite{wang2018learning}, RNN-HA~\cite{xiu2018coarse} and RAM~\cite{liu2018ram} which we have introduced in detail. 
Here we analyze the experimental results in this subsection.

The RNN-HA~\cite{xiu2018coarse} only used ID classification loss and the vehicle type annotation during the training stage. 
Its performance is slightly better than \textbf{ID Classification Loss} and lower than {\textbf{Triplet$+$ID Loss}}.

Both the MGN~\cite{wang2018learning} and RAM~\cite{liu2018ram} split the image horizontally to extract the local feature. 
But there are wide range of view-angle changes in our datasets as the images are captured by UAVs. 
The two approaches cannot be simply transferred to our task, because it is not easy to align corresponding parts if they only split the image horizontally. 
Thus, these two algorithms do not achieve good performance in our datasets.

For our method, we use triplet loss to improve the model, and we use many kinds of attributes classification loss to improve the performance of our algorithm. 
What's more, we have also focused on extracting the local feature by using the discriminative parts detector and weighted pooling. 
As shown in Table~\ref{tab:results} our algorithm achieves a better performance than the other three state-of-art methods.

\subsection{Human Performance Evaluation}
In order to investigate the difficulty of our dataset and the performance gap between human and our algorithm, 
we conducted human performance evaluation on our dataset. 
In this experiment, we randomly select $100$ query images, coming with two candidates to be matched. 
One candidate has the same ID to the query, and the other is selected from other IDs but with the same annotated attributes.   
Three well-trained annotators are involved in this experiment, spending $30$ seconds for one query in average.
The performances of these annotators are shown in Table~\ref{tab:hu}.

Figure~\ref{fig:hum} shows some examples of the human performance evaluation experiment. Query images, annotators' results and our algorithm's results are listed on the left, middle, and right column, respectively. It can be seen that vehicle poses and camera view-angles are important factors affecting the algorithm performance.
By comparison, human performance is more insensitive to these factors.

\section{Conclusion}\label{sec_conc}
In this paper, we collect the VRAI dataset which is the largest aerial vehicle ReID dataset so far to our knowledge. Besides the identity, we also provide additional annotation informations such as the color, vehicle type, attributes \eg whether it contains skylight, spare tire \etc and discriminative parts of the images in the dataset. So our dataset can be used in many other vision tasks such as fine-grained classification and attribute prediction.
In addition, it is worth noting that the view-angle of the vehicles is diverse in single UAV platform, not to mention we have two UAVs which are controlled by different pilots to fly in different locations.
Furthermore, we also conducted comprehensive experiments to take the full advantage of the rich annotations. Based on the rich annotation information of our dataset, we propose a novel approach for vehicle ReID from aerial images, which is capable of explicitly detecting discriminative parts for each specific vehicle and significantly outperform three promising baseline methods and three other ReID methods evaluated on our dataset.
In Figure~\ref{fig:Overview}, the image-pairs of the same vehicles are (1,2), (1,3) and (1,3) for Row 1, 2 and 3, respectively.

{\small
\bibliographystyle{ieee}
\bibliography{egbib}
}

\end{document}